\documentclass[conference]{IEEEtran}
\IEEEoverridecommandlockouts
\usepackage{cite}
\usepackage{amsmath,amssymb,amsfonts}
\usepackage{algorithmic}
\usepackage{graphicx}
\usepackage{textcomp}
\usepackage{xcolor}
\usepackage{multirow}
\usepackage{amsmath} 
\usepackage{booktabs}
\usepackage[hang,small,bf]{caption}
\usepackage[subrefformat=parens]{subcaption}
\def\BibTeX{{\rm B\kern-.05em{\sc i\kern-.025em b}\kern-.08em
    T\kern-.1667em\lower.7ex\hbox{E}\kern-.125emX}}
\begin{document}

\title{SETN: Stock Embedding Enhanced with Textual and Network Information

}

\author{\IEEEauthorblockN{Takehiro Takayanagi}
\IEEEauthorblockA{\textit{The University of Tokyo} \\
\textit{School of Engineering}\\
Tokyo, Japan \\
m2021ttakayanagi@socsim.org}
\and
\IEEEauthorblockN{ Hiroki Sakaji}
\IEEEauthorblockA{\textit{The University of Tokyo} \\
\textit{School of Engineering}\\
Tokyo, Japan }
\and
\IEEEauthorblockN{Kiyoshi Izumi}
\IEEEauthorblockA{\textit{TheUniversity of Tokyo} \\
\textit{School of Engineering}\\
Tokyo, Japan}
}

\maketitle

\begin{abstract}
    Stock embedding is a method for vector representation of stocks.
    There is a growing demand for vector representations of stock, i.e., stock embedding, in wealth management sectors, and the method has been applied to various tasks such as stock price prediction, portfolio optimization, and similar fund identifications. 
    Stock embeddings have the advantage of enabling the quantification of relative relationships between stocks, and they can extract useful information from unstructured data such as text and network data.
    In this study, we propose stock embedding enhanced with textual and network information (SETN) using a domain-adaptive pre-trained transformer-based model to embed textual information and a graph neural network model to grasp network information.
    We evaluate the performance of our proposed model on related company information extraction tasks.
    We also demonstrate that stock embeddings obtained from the proposed model perform better in creating thematic funds than those obtained from baseline methods, providing a promising pathway for various applications in the wealth management industry.
\end{abstract}

\begin{IEEEkeywords}
Stock embedding, Stock analysis, Natural language processing, Graph embedding, Knowledge graph, Graph neural network
\end{IEEEkeywords}

\section{Introduction}
Financial experts, such as fund managers, financial advisors, and traders, utilize the relative relationships between stocks, as well as individual stock characteristics, for their professional activities.
For example, traders choose stock pairs with high positive correlations based on financial time series in pairs trading, and fund managers select stocks with similar themes to create thematic mutual funds.
Therefore, it is important to quantify the relative relationships between stocks for quantitative stock analysis.

One method of capturing relative relationships between stocks is stock embedding.
Stock embedding is a powerful tool to extract relative relationships between stocks from various data by expressing stocks with vector representations.
Stock embedding has two advantages over existing stock analysis methods. 
First, it enables the quantification of relative relationships by calculating cosine similarities. 
Second, unstructured data, such as textual and network data, can be utilized through various embedding methodologies\cite{Kipf2017,velikovi2017,Grover2016,Devlin2019,Liu2019}.

Financial text data such as annual reports, integrated reports, and earnings summaries are often utilized to obtain stock embeddings\cite{Ito2020, Du2020}.
Textual data are especially useful in capturing individual stock information, as they contain a large amount of this information, including prevailing business situations, management discussions, and current business risks.
Several text embedding methods have been employed to obtain stock embeddings from textual data\cite{Jeffrey2014,Vaswani2017,Cer2018}.
Studies have shown that pretrained transformer-based models effectively learn stock embeddings from textual data\cite{Ito2020}.

Similarly, network data such as supplier-customer relationships, transaction networks, and fund-asset relationships are used to gain knowledge of stock embeddings\cite{Wu2021,Satone2021,Li2020,Li2019 ,Chen2018_2,Ye2021,Matsunaga2019,Gao2021}.
Network data are useful in capturing cross-stock information, as graphs help represent the complex nature of financial systems with many components and sophisticated relations.
Several graph embedding methods have been employed to learn stock embeddings from network data\cite{Grover2016,Kipf2017,velikovi2017}. 
Studies have found that graph neural network (GNN) models are effective in learning stock embeddings from network data\cite{Chen2018_2,Ye2021,Gao2021,Matsunaga2019}.

Ideally, stock embeddings should represent stocks from various perspectives. 
However, many studies do not consider it necessary to train models with both textual and network data, although both types of data have been shown to be effective for stock embedding \cite{Chen2018_1, Ito2020, Du2020, Wu2021 ,Li2019 ,Li2020, Satone2021 ,Chen2018_2, Ye2021 ,Gao2021,Matsunaga2019}. 
Therefore, it is worth exploring a model that can jointly learn textual and network data. 

In this study, we propose SETN, which jointly trains a transformer-based and GNN model. 
First, SETN further pretrains a pretrained transformer-based model on a financial corpus to adapt the model to the financial domain\cite{Liu2020,Peng2021,Araci2019}.
Second, it learns textual information from the annual reports of a target company using a further pre-trained transformer-based model. 
It then creates a subgraph of the target stock from the network data and learns the network information using a GNN.
Finally, we added a single-layer feedforward neural network to predict the sector and industry of a target stock. 

We conducted experiments on the Japanese stock market to evaluate the performance of our proposed model.
Our SETN model outperforms the baseline models in related company information extraction tasks, in which  stocks in similar business sectors and industries are extracted.
The results suggest the benefit of jointly learning both textual and network information.
We also analyze the effect of graph type on performance, as there has been limited analysis of what type of graph better captures network information.
Furthermore, we investigate the effectiveness of jointly training a transformer-based and GNN models by separately comparing them with a transformer-based and GNN model. 
The results suggest that using a directed graph leads to higher performance, and that joint training contributes to higher performance. 
Furthermore, we introduce thematic fund creation tasks, in which we extract stocks with similar themes.
We show that embeddings obtained using our proposed model extract themed stock better than embeddings from the baseline models.  

The main contributions of our study are as follows: 
\begin{enumerate}
    \item We propose SETN. The results of experiments show our method outperforms baseline models on related company information extraction tasks. 
    \item  We conduct a comparative study of graph types and learning architectures in stock embeddings.
    \item We show that the stock embeddings obtained from SETN can be applied to the creation of thematic funds by introducing thematic fund creation tasks. 
\end{enumerate}

\section{Related Work}
Stock embedding is a powerful tool used for vector representation of various aspects of stocks. 
It has been applied to multiple tasks, such as stock price prediction, portfolio optimization, and similar fund identification\cite{Chen2018_1, Ito2020, Du2020, Wu2021 ,Li2019 ,Li2020, Satone2021 ,Chen2018_2, Ye2021 ,Gao2021,Matsunaga2019}. 

Stock embedding methods can be classified into two approaches based on the properties of the data used for embeddings.
The first approach uses individual stock information\cite{Du2020, Ito2020,Chen2018_1}.
Traditionally, numerical data, such as financial time series and technical indicators, have been used to obtain latent representations of the stock market.
Chen {\it et al.} \cite{Chen2018_1} extracted stock market embeddings from a financial time series using an attention-based long short-term memory (LSTM) model to predict the daily return ratio.
Recently, textual data, such as annual reports and earnings summaries, have been utilized to obtain stock embeddings as textual data contain rich individual stock information.
Transformer-based models effectively learn individual stock information from textual data\cite{Ito2020, Du2020}.
Ito {\it et al.} \cite{Ito2020} extracted company stock embeddings from a company’s annual reports using bidirectional encoder representations from transformers (BERT) to obtain fine-grained industry information.
Du {\it et al.} \cite{Du2020} extracted stock embeddings from news articles using BERT and achieved portfolio optimization.

The second approach utilizes cross-stock information\cite{Chen2018_2,Ye2021, Wu2021,Gao2021,Satone2021,Li2020,Li2019 ,Matsunaga2019}.
This approach utilizes various network data to embed the topological information of the stocks.
The network data used for stock embeddings include shareholding graphs\cite{Chen2018_2,Ye2021}, news co-occurrence graphs\cite{Wu2021}, fund-asset relation graphs\cite{Satone2021,Li2020,Li2019}, and supplier-customer relation graphs\cite{Matsunaga2019}.
Several graph embedding methods have been proposed for stock embedding.
Wu {\it et al.} \cite{Wu2021} obtained stock embeddings using matrix decomposition of news co-occurrence graphs for stock price predictions.
Several researchers, such as \cite{Satone2021,Li2020,Li2019} obtained stock embeddings by using the node2vec algorithm\cite{Grover2016} on fund-asset graphs to identify similar funds and obtain new technical indicators.
Recently, the GNN method has evolved and has been shown to be effective in extracting stock embeddings from cross-stock relationships\cite{Chen2018_2,Ye2021,Gao2021,Matsunaga2019}.
\begin{table}[]
    \centering
    \caption{TOPIX17\&TOPIX33, Sector and Industry Classification in Japanese Stock Market}
    \label{tab:table1}
    \scriptsize
    \scalebox{0.8}[1.0]{
        \begin{tabular}{ccl}
        \hline
        TOPIX17                                                                                                 & TOPIX33                              &  \\ \hline
        \multirow{2}{*}{FOODS}                                                                                   & Fishery, Agriculture \& Forestry      &  \\ \cline{2-3} 
                                                                                                                 & Foods                                 &  \\ \hline
        \multirow{2}{*}{ENERGY RESOURCES}                                                                        & Mining                                &  \\ \cline{2-3} 
                                                                                                                 & Oil and Coal Products                 &  \\ \hline
        \multirow{3}{*}{CONSTRUCTION\&MATERIALS}                                                                 & Construction                          &  \\ \cline{2-3} 
                                                                                                                 & Metal Products                        &  \\ \cline{2-3} 
                                                                                                                 & Glass and Ceramics Products           &  \\ \hline
        \multirow{3}{*}{RAW MATERIALS\&CHEMICALS}                                                                & Textiles and Apparels                 &  \\ \cline{2-3} 
                                                                                                                 & Pulp and Paper                        &  \\ \cline{2-3} 
                                                                                                                 & Chemicals                             &  \\ \hline
        PHAMACEUTICAL                                                                                            & Pharmaceutical                        &  \\ \hline
        \multirow{2}{*}{\begin{tabular}[c]{@{}c@{}}AUTOMOBILES\&TRANSPORTATION \\ EQUIPMENT\end{tabular}}        & Rubber Products                       &  \\ \cline{2-3} 
                                                                                                                 & Transportation Equipment              &  \\ \hline
        \multirow{2}{*}{STEEL\&NONFERROUS METALS}                                                                & Iron and Steel                        &  \\ \cline{2-3} 
                                                                                                                 & Nonferrous Metals                     &  \\ \hline
        MACHINERY                                                                                                & Machinery                             &  \\ \hline
        \multirow{2}{*}{\begin{tabular}[c]{@{}c@{}}ELECTRIC APPLIANCES\&PRECISION   \\ INSTRUMENTS\end{tabular}} & Electric Appliances                   &  \\ \cline{2-3} 
                                                                                                                 & Precision Instruments                 &  \\ \hline
        \multirow{3}{*}{IT\&SERVICES, OTHERS}                                                                    & Other Products                        &  \\ \cline{2-3} 
                                                                                                                 & Information \& Communication          &  \\ \cline{2-3} 
                                                                                                                 & Services                              &  \\ \hline
        ELECTRIC POWERT\&GAS                                                                                     & Electric Power and Gas                &  \\ \hline
        \multirow{4}{*}{TRANSPORTATION\&LOGISTICS}                                                                & Land Transportation                   &  \\ \cline{2-3} 
                                                                                                                 & Marine Transportation                 &  \\ \cline{2-3} 
                                                                                                                 & Air Transportation                    &  \\ \cline{2-3} 
                                                                                                                 & Warehousing and Harbor Transportation &  \\ \hline
        COMMERCIAL\&WHOLESALE TRADE                                                                              & Wholesale Trade                       &  \\ \hline
        RETAIL TRADE                                                                                             & Retail Trade                          &  \\ \hline
        BANKS                                                                                                    & Banks                                 &  \\ \hline
        \multirow{3}{*}{FINANCIAL(EXCEPT BANKS)}                                                                     & Securities and Commodities Futures    &  \\ \cline{2-3} 
                                                                                                                 & Insurance                             &  \\ \cline{2-3} 
                                                                                                                 & Other Financing Business              &  \\ \hline
        REAL ESTATE                                                                                              & Real Estate                           &  \\ \hline
        \end{tabular}}
        \end{table}

Several studies have been conducted on the extraction of  stock embeddings using either textual or network information, which have been shown to be useful for extracting stock embeddings\cite{Chen2018_1, Ito2020, Du2020, Wu2021 ,Li2019 ,Li2020, Satone2021 ,Chen2018_2, Ye2021 ,Gao2021,Matsunaga2019}.
However, notably, 
the number of studies that focus on using both textual and network information is quite limited.
Some studies such as \cite{Sawhney2020_1,Sawhney2020_2} used both types of information by encoding textual information with a transformer-based model and using them as input to a GNN model.
These studies differ from ours in two ways.
First, they did not jointly train text embedding and graph embedding models.
Specifically, they do not update the parameters for text embedding using existing pre-trained transformer-based models\cite{Cer2018}, whereas our approach fine-tunes a transformer-based model. 
Second, their study aimed to predict economic indicators and stock prices, while our study aimed to extract stock embeddings.
Therefore, the effectiveness of learning stock embeddings from both textual and network data merits further investigation.

\section{Methodology}
\begin{figure*}
    \centering
        \includegraphics[width=\linewidth]{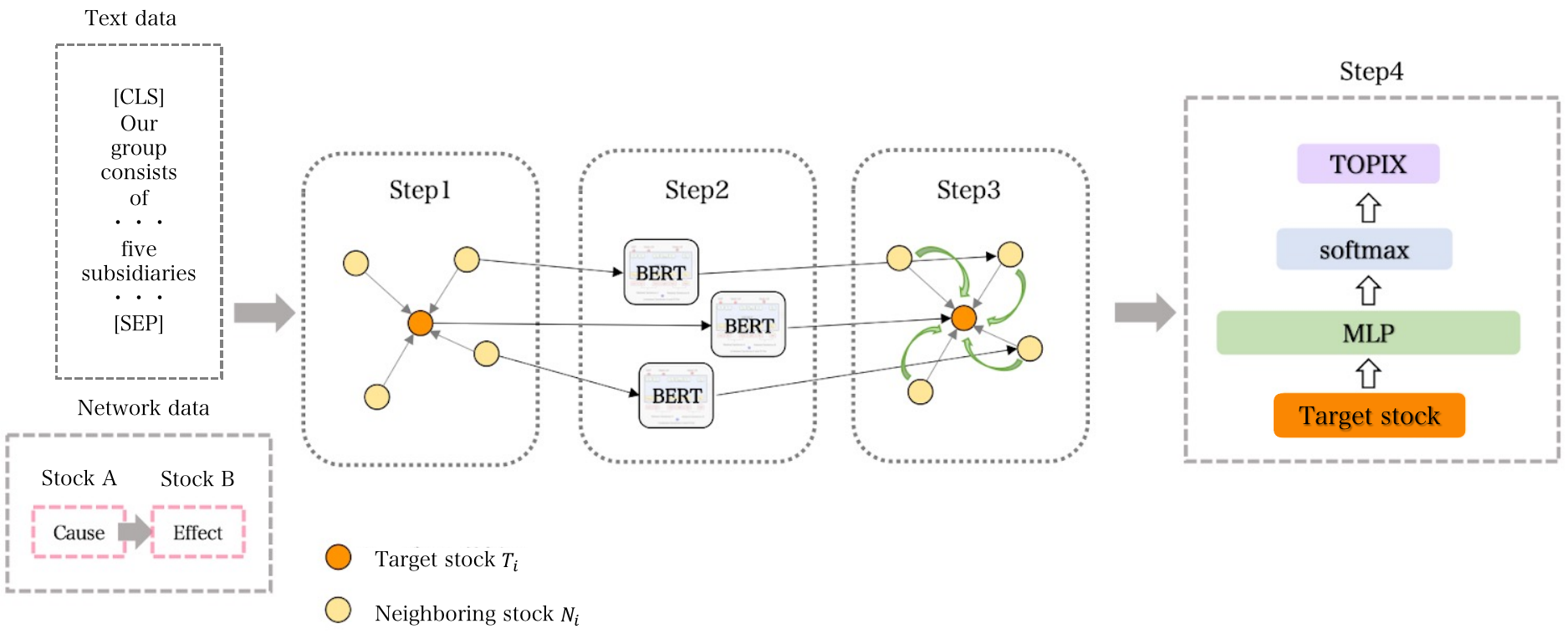}
        \caption{Outline of our proposed model SETN}
        \label{fig:figure1}
    
\end{figure*}
In this section, we provide an overview of SETN, a stock embedding enhanced with textual information and network information that jointly trains a transformer-based and GNN model, followed by a detailed explanation of each component.
As shown in Figure \ref{fig:figure1}, the model approach consists of the following four steps.
Step 1 extracts a subgraph centered on a target node using network data. 
Step 2 obtains embeddings for each node from textual information usin the domain-adaptive pretrained transformer-based model.
Step 3 captures the network information using the GNN model.
Step 4 feeds the embeddings of the target stock into the classifier for the sector and industry classifications listed in Table \ref{tab:table1}.

\subsection{Transformer-based Model for Text Embedding}
We extracted textual information by embedding the business description of a company using a transformer-based model to obtain a sequence of token embeddings.
We can either use BERT\cite{Devlin2019} or a robust optimized BERT pre-training approach (RoBERTa)\cite{Liu2019}.
Before applying pre-trained BERT or RoBERTa to SETN, we implemented domain-adaptive pre-training using financial corpora, as several studies have suggested adapting the BERT architecture to a specific domain, such as medicine and finance, has a significant effect on model performance\cite{Liu2020,Peng2021,Araci2019}.
Subsequently, we applied pooling operations to the token embeddings to obtain fixed-size embeddings. 
Three pooling strategies are possible: (1) taking the output of the special classification (CLS) token, (2) taking the mean of all the output vectors, and (3) applying max-pooling operations over the output vectors. 
SETN employs the second pooling strategy, taking the mean of all output vectors, because the strategy has been proven to be most effective for semantic textual similarity, natural language inference, and sector and industry classification\cite{Reimers2019,Ito2020}.

\begin{equation}
    h_{\mathrm{BERT},i} = \mathrm{BERT}(x_i)
\end{equation}
where $x_i$ is the business description of $\mathrm{target\,stock}\:i$ and $h_{\mathrm{BERT},i}\in{\mathbb{R}}$ is the pooled output of the BERT. 

\subsection{GNN Model for Graph Embedding}
We extracted network information using the GNN.  
GNN has been gaining popularity because it enables the capturing of structural relationships by updating and aggregating node representations in the financial domain\cite{Wang2021}.
The GNN models that we employed in the experiments were the graph convolutional network (GCN)\cite{Kipf2017} and graph attention network (GAT)\cite{velikovi2017}.
The GCN is one of the most widely used GNN models that aggregates features of neighborhood nodes and normalizes the aggregated representations by node degree.

\begin{equation}
    h_{\mathrm{GCN}, i} = \mathrm{ReLU}\left(\sum_{j\in{N_i}}\hat{D}^{-\frac{1}{2}}\hat{A}\hat{D}^{-\frac{1}{2}}h_jW+b\right)
\end{equation}
where $N_i$ is the neighboring stock of $\mathrm{target\,stock}\:i$,
$\hat{A}=A+I$ denotes the adjacency matrix of the graph with inserted self-loops, and $\hat{D}_{i,i}=\sum_{j=0}\hat{A}_{ij}$ is the diagonal degree matrix.
$h_i$ is the input for the GCN, $W$ is the weight matrix, and $b$ is the bias.

The GAT assigns neighboring nodes with different weight importance by incorporating an attention mechanism during aggregation.
\begin{equation}
    a_{ij}=\frac{\exp\left(\mathrm{LeakyReLU}\left(a\left[Wh_i\|Wh_j\right]\right)\right)}
    {\sum_{k\in{N_i}}\exp\left(\mathrm{LeakyReLU}\left(a\left[Wh_i\|Wh_k\right]\right)\right)} 
\end{equation}
where $a$ is an attention mechanism, and $\|$ represents a concatenation operation. We adopted the leaky ReLU (LReLU or LReL) function with a $negative slope=0.2$ for our activation function.

\subsection{Residual Connection}
Research shows that adding a residual connection (RC) to a GNN model improves its performance \cite{Liu2021}.
Thus, we employed a residual connection by adding the output of the BERT to the output of the GNN. 
The residual connection is expressed as follows:
\begin{equation}
    \begin{split}
      &h_{\mathrm{BERT},i} = \mathrm{BERT}(x_i)
      \\
      &h_{\mathrm{GCN},i} = \mathrm{GCN}(h_{\mathrm{BERT},i})
      \\
      &h_{\mathrm{res},i} = h_{\mathrm{BERT},i}+h_{\mathrm{GCN},i}
    \end{split}
    \end{equation}

\subsection{Loss Function}
In this study, we considered a multitask learning setup for sector and industry classification tasks. For this task, we used the Tokyo Price Index-17 (TOPIX17) labels and the Tokyo Price Index-33 (TOPIX33) labels defined by the Tokyo Stock Exchange (TSE) in Table \ref{tab:table1}.
TOPIX33 is an industrial sector index traditionally used in the Japanese market while TOPIX33 reorganizes itself to TOPIX17 for convenience.   
This loss was due to the addition of TOPIX17 loss and TOPIX33 loss.
\begin{equation}
    Loss = Loss^{\mathrm{TOPIX17}}+Loss^{\mathrm{TOPIX33}}
\end{equation}
The TOPIX17 loss and TOPIX33 loss were computed with a classifier of a single-layer feedforward neural network with dropout and ReLU activation. 
\begin{equation}\label{eq:loss}
    \begin{split}
      &y_i^{\mathrm{TOPIX}}=\mathrm{Softmax}\left(W^{\mathrm{TOPIX}}h_i+b^{\mathrm{TOPIX}}\right)
      \\
      &Loss^{\mathrm{TOPIX}}=\sum_{i\in{\Omega}}\mathrm{CE}\left(d^{\mathrm{TOPIX}}_i, y^{\mathrm{TOPIX}}_i\right)
    \end{split}
\end{equation}
where $h_i$ represents the output of the GNN or a residual block, $d_i$ represents the target label of $\mathrm{target\,stock}\:i$, $\Omega$ represents all stocks, and
CE represents the cross-entropy loss function. In Equation \ref{eq:loss}, we denote TOPIX to show either TOPIX17 or TOPIX33, because we calculate the loss of TOPIX17 and 
TOPIX33 in a similar manner.

\section{Dataset}
We conducted our experiments on the Japanese stock market. 
For textual information, we used business descriptions from companies' annual reports in Japanese language datasets on the Japanese stock market.
For network information, we used causal chains\cite{Izumi2019}, which represent cross-company economic relationships. 
Overall, our dataset contained 2,437 Japanese stocks. We split the dataset into training, validation, and test data containing respectively 1,705, 243, and 489 companies.

\subsection{Text Corpora}
We used CoARiJ\footnote{https://github.com/chakki-works/CoARiJ}, The Corpus of Annual Reports in Japan, which is organized by Japanese financial reports. 
The corpus covers 3,016 companies and includes annual reports, integrated reports, corporate social responsibility (CSR) reports, and earnings summaries, ranging from 2014 to 2018. 
We used all corpora for domain-adaptive pre-training. 
In conducting the experiments, to embed textual information, usef a company's business description, which contains an overview of the activities of the company.

\subsection{Network Data}
In this study, we employed a causal chain as company network data.
The causal chain concept was proposed by Izumi {\it et al.} \cite{Izumi2019} and is composed of causality in textual data.
For example, the sentence ``U.S. stocks give up earlier gains as oil recovers'' includes causality that is composed of a cause ``oil recovers'' and an effect ``U.S. stocks give up earlier gains.''
They detect causality using clue expressions such as ``as.''
The meanings of the cause, effect, and clue expressions are defined in \cite{Izumi2019}.
They then collect cause and effect pairs from financial reports using this technique.
Next, they connect the cause to the effect based on cosine similarities from word2vec.
For example, the sentence ``World oil prices hit a three- and half-month high on Tuesday after militants said they would broaden attacks on Nigeria's oil industry'' includes a cause ``militants said they would broaden attacks on Nigeria's oil industry,'' and an effect ``World oil prices hit a three and half-month high on Tuesday.''
The effect can be connected to the previous cause ``oil recovers”; then, we can obtain the causal chain ``militants said they would broaden attacks on Nigeria's oil industry'' -> ``World oil prices hit a three and half-month high on Tuesday'' and ``oil recovers'' -> ``U.S. stocks give up earlier gains as oil recovers.''
Izumi {\it et al.} constructed causal chains using Japanese financial reports and accessed the causal chain search system on their website \footnote{https://socsim.t.u-tokyo.ac.jp/ccs/}.
Moreover, their causal chains are linked to each company because they use the financial reports of each company.
Therefore, causal chains can be used as network data for companies.
Furthermore, several studies have suggested the effectiveness of representing cross-company economic relationships using causal chains\cite{Nakagawa2019, Nakagawa2022}.
In our study, a company is represented as a node, and a company’s relationship based on the causal chain is represented as an edge. The network has 2,437 nodes and 12,755 edges.


\subsection{Thematic Fund}
For the thematic fund creation tasks, we utilized themed stock listed on a website called ``minkabu,''\footnote{https://minkabu.jp/screening/theme} a Japanese financial information services website.
We extracted 27 themes using the criterion that a theme should include more than 15 stocks in our test data. 
For the thematic fund creation task, we selected six themes: logistics, IT, biotechnology, semiconductor, 5G and Post-COVID, and assigned them to one of three separate groups. 
Logistics and IT belong to the first group. The characteristics of the first group are that the theme is related to the sector and industry, and it has a similar label in TOPIX17 and TOPIX33. 
Biotechnology and semiconductors are in the second group. The characteristic of the second group is that the theme is related to sector and industry, but it does not have similar labels in TOPIX17 or TOPIX33. 
The third group included the 5G and post-COVID. For the third group, the theme is not about either sector or industry, but other concepts such as social issues and new technology.

\begin{figure*}[htbp]
    \begin{minipage}[b]{0.35\linewidth}
      \centering
      \includegraphics[width=\linewidth]{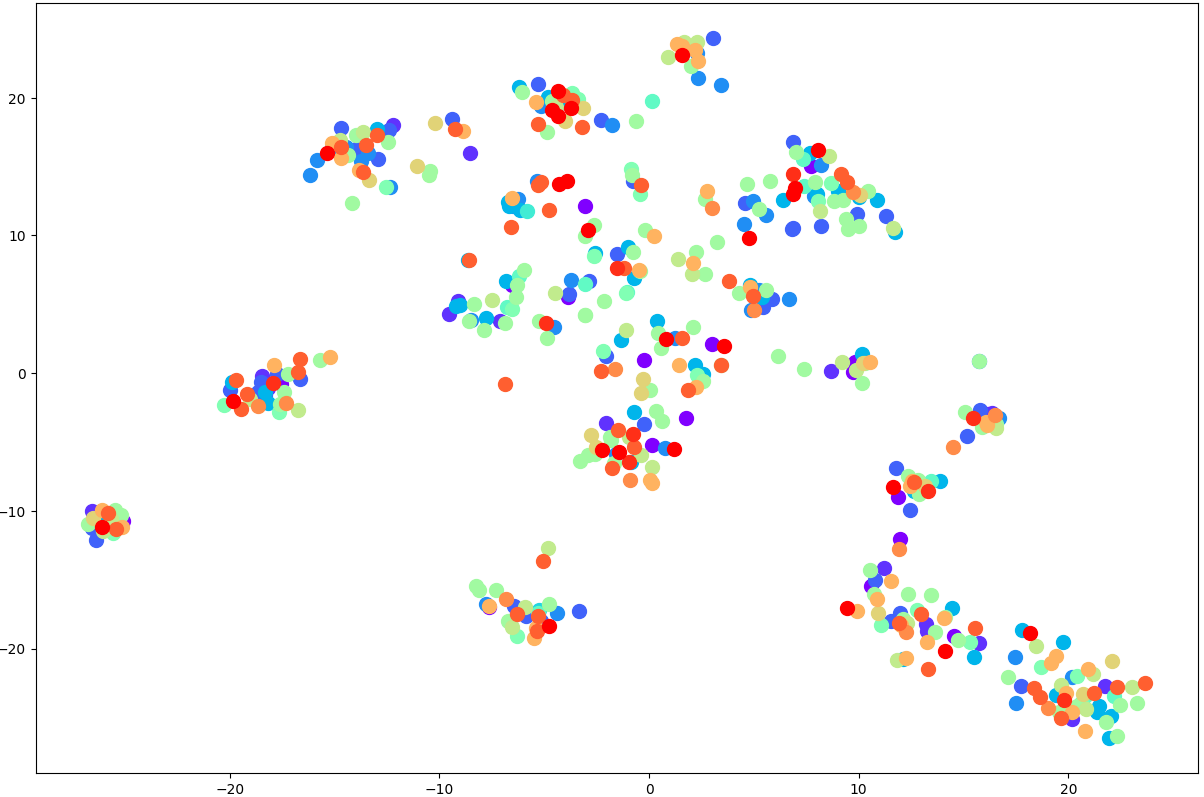}
      \subcaption{BERT}
      \label{fig:figure2a}
    \end{minipage}
    \begin{minipage}[b]{0.35\linewidth}
      \centering
      \includegraphics[width=\linewidth]{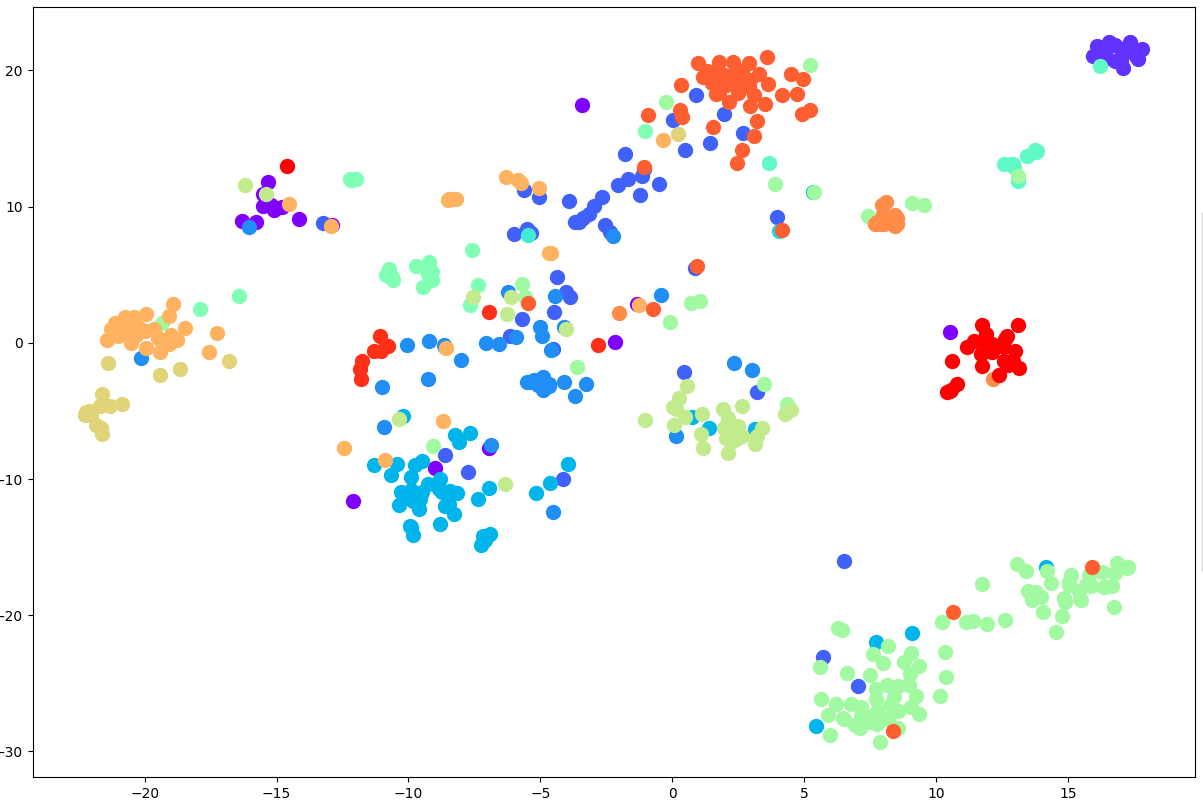}
      \subcaption{BERT+GCN+RC}
      \label{fig:figure2b}
    \end{minipage}
    \begin{minipage}[b]{0.3\linewidth}
    \centering
    \includegraphics[width=\linewidth]{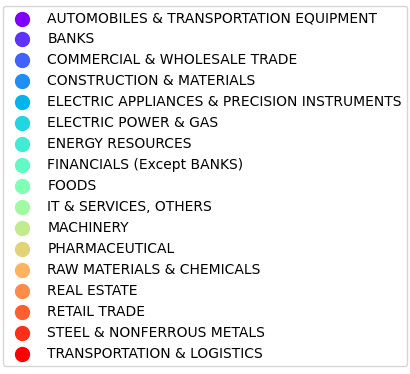}
    \end{minipage}
    \caption{Visualization of the embeddings with TOPIX17 sector labels.
    Subfigure (a) shows the embeddings from the BERT model. Subfigure (b)
    shows the embeddings from the BERT+GCN+RC model}
    \label{fig:figure2}
\end{figure*}

\begin{table*}[]
    \centering
    \caption{Results for Related Company Information Extraction}
    \label{tab:table2}
    \begin{tabular}{lccccccc}

    \toprule
                   &                & TOPIX17       &                &           &                & TOPIX33       &                \\
    \cmidrule(lr){2-4} 
    \cmidrule(lr){6-8} 
                   & MAP@5          & MAP@10         & MAP@50         &           & MAP@5          & MAP@10         & MAP@50         \\ 
    BERT           & 0.669          & 0.628          & 0.513          &           & 0.601          & 0.572          & 0.508          \\
    RoBERTa        & 0.669          & 0.631          & 0.532          &           & 0.581          & 0.543          & 0.488          \\
    \midrule
    BERT+GCN       & 0.635          & 0.594          & 0.484          &           & 0.551          & 0.516          & 0.455          \\
    BERT+GAT       & 0.635          & 0.588          & 0.467          &           & 0.555          & 0.517          & 0.436          \\
    RoBERTa+GCN    & 0.632          & 0.598          & 0.5            &           & 0.548          & 0.516          & 0.451          \\
    RoBERTa+GAT    & 0.616          & 0.576          & 0.474          &           & 0.527          & 0.496          & 0.44           \\
    \midrule
    BERT+GCN+RC    & ${\bf 0.682}$ & 0.651          & 0.547          &           & ${\bf0.606}$ & 0.587          & ${\bf0.528}$ \\
    BERT+GAT+RC    & 0.679          & ${\bf0.657}$ & ${\bf0.561}$ &           & 0.598          & ${\bf0.585}$ & 0.524          \\
    RoBERTa+GCN+RC & 0.68           & 0.651          & 0.553          &           & 0.599          & 0.579          & ${\bf0.528}$ \\
    RoBERTa+GAT+RC & 0. 666         & 0.635          & 0.528          &           & 0.581          & 0.555          & 0.495      \\
    \bottomrule   

\end{tabular}
\end{table*}

\section{Experiment}
\subsection{Training Details}

We leveraged the pre-trained Japanese BERT and RoBERTa models for our proposed model.
We used the following pre-trained models from Hugging Face: bert-base-Japanese \footnote{https://huggingface.co/cl-tohoku/bert-base-japanese}, and roberta-base-Japanese \footnote{https://huggingface.co/nlp-waseda/roberta-base-japanese}.
The models were trained on Japanese text with 12 layers, 768 hidden units, and 12 attention heads.
The text was tokenized by MeCab\footnote{http://mecab.sourceforge.jp} and WordPiece\cite{Schuster2012} for bert-base-japanese, 
The text was first segmented into words by Juman++\cite{Arseny2018} in advance, and each word was tokenized by SentencePiece\cite{Kudo2018} for roberta-base-Japanese.

For domain-adaptive pre-training, we used all text data from the CoARiJ corpus (approximately 200,000 text files and 5.0 GB).
The pre-training was based on the masked language model and next-sentence prediction for the BERT model and masked language model for RoBERTa. 
We trained all 12 layers in 200,000 steps in both models.

For our proposed model, we utilized the first 512 tokens of business descriptions in annual reports as textual information.
In training SETN, we trained only the last layer in BERT or RoBERTa.
For extracting a subgraph, we performed neighborhood sampling by one hop in this study.
We backpropagated only through the loss of the target node in the subgraph to avoid leakage.
We trained our models on the loss function using the Adam optimizer\cite{Kingma2014} for 20 epochs, with a learning rate of 0.001 and dropout rate of 0.2. 
For all models, PyTorch and PyTorch-geometric were used for implementation. 
All experiments were conducted on an NVIDIA A6000 and Intel i9-10980XE.

\subsection{Evaluation Tasks}
First, we evaluated the performance of our proposed method on a related company information extraction task.
We then conducted two types of ablation studies.
Finally, we evaluated the performance of our proposed model on a thematic fund creation task.

\subsection*{Related company information extraction task}
The purpose of the related company information extraction task is to evaluate the extent to which stock embeddings capture sector and industry information.
Following the work in\cite{Yu2012}, we evaluated the embeddings using the Mean Average Precision at K (MAP@K) metric.
We set K in MAP@K to $K=5,10,50$.

To verify the usefulness of our proposed model further, we conducted two types of ablation studies.
The first ablation study analyzed the effect of the graph type on performance.
Most studies use undirected graphs for graph construction\cite{Chen2018_2,Gao2021,Wu2021,Satone2021,Li2020} although the relationships between stocks are often described in a directed form, such as supply chain and transaction relationships. 
Therefore, we compared the performance of the models with a directed graph and evaluated the model's performance with an undirected graph.
The second ablation study verified the improvement in model performance by jointly training a text embedding and graph embedding model, compared to several studies in which text embedding and graph embedding models were trained  separately\cite{Sawhney2020_1, Sawhney2020_2}.
Specifically, we compared our proposed model with a model that does not update the parameters of a transformer-based model. In other words, the text features are encoded via a transformer-based model and fed as an input to a graph embedding model.

\subsection*{Thematic fund creation tasks}
In this task, we evaluated the model's capability to find relevant companies with respect to a given theme.
We experimented with thematic fund creation tasks based on six different themes.
To this end, we obtained ${|\bf t|}$ most similar companies for each $\mathrm{stock}\:i$ in a given theme in terms of cosine similarities. 
Then, we evaluated how many stocks out of ${|\bf t|}$ stocks shared the same theme as $\mathrm{stock}\:i$ using the following evaluation metric.

\begin{equation}
    \text{theme 
    metric}=\sum_{{\bf t}\in\mathcal{T}}\sum_{i,j=1}^{|\bf t|} \frac{\mathcal{H}_{i,j,{\bf t}}}
    {|\mathcal{T}| |{\bf t}|^2} 
\end{equation}

where $\mathcal{T}$ is a set of all themes and  ${\bf t}$ represents a set of stocks in a given theme, such as biotechnology and post-COVID.
$\mathcal{H}_{i,j,{\bf t}}$ are calculated as follows: 

\begin{equation}
    \mathcal{H}_{i,j,{\bf t}}=\\
    \left\{
    \begin{array}{@{}ll@{}}
      1, & \text{{\small if $pred_{i,j}$ is in the same theme as stock\:$i$}} \\
      0, & \text{{\small otherwise}}
    \end{array}\right. 
\end{equation} 

where $pred_{i,j}$ is the stock closest to $\mathrm{stock}\:i$ with respect to cosine similarity.

\subsection{Baseline Models}
We compared the performance of the proposed model with the following baselines: 
\subsubsection*{BERT$_{Japanese}$}
Pre-trained BERT with single-layer feedforward neural network. The pre-trained BERT was pre-trained on Japanese Wikipedia and further pre-trained on the CoARiJ dataset. 
\subsubsection*{RoBERTa$_{Japanese}$}
RoBERTa was pre-trained with a single-layer feedforward neural network. The pre-trained RoBERTa was pre-trained on Japanese Wikipedia and the Japanese portion of CC-100 and further pre-trained on the CoARiJ dataset.

\subsection{Result}

\begin{table*}[]
    \scriptsize
    \centering
    \caption{Three nearest neighbors for selected stocks in the test set for the BERT model}
    \label{tab:table3}
    \begin{tabular}{ccc}
        
                                       \toprule
                                       Company                            & TOPIX17                                       & TOPIX33                              \\
    \midrule
    \midrule
    Kintetsu Group   Holdings Co., Ltd.  & Transportation \& Logistics               & Land Transportation                   \\
    \midrule
    Tobu Railway Co., Ltd.               & Transportation \& Logistics               & Land Transportation                   \\
    Takashimaya Co., Ltd.       & Retail Trade                               & Retail Trade                        \\
    Kintetsu Department Store Co., Ltd.  & Retail Trade                               & Retail Trade                        \\
    \midrule
    \midrule
    Nippon Express Co., Ltd.          & Transportation \& Logistics               & Land Transportation                   \\
    \midrule
    Kintetsu World Express, Inc        & Transportation \& Logistics               & Warehousing and Harbor Transportation\\
    Niigata Kotsu Co., Ltd.              & Transportation \& Logistics               & Land Transportation                   \\
    Kyushu Railway Company             & Transportation \& Logistics               & Land Transportation                   \\
    \midrule
    \midrule
    Fujitsu General Limited            & Electric appliances \& Precision instruments & Electric Appliances                   \\
    \midrule
    Komori Machinery (Nantong) Co., Ltd & Machinery                                   & Machinery                             \\
    Towa Corporation                  & Machinery                                   & Machinery                             \\
    Shibaura Mechatron Corporation  & Electric appliances \& Precision instruments & Electric Appliances                   \\
    \midrule
    \midrule
    SBI Holdings, Inc                  & Financial (except banks)                        & Securities and Commodities Futures    \\
    \midrule
    Yodobashi Holdings INC            & Retail Trade                               & Retail Trade                        \\
    Hotman Co., Ltd.                    & Retail Trade                               & Retail Trade                        \\
    Square Enix Holdings Co., Ltd.      & IT \& Services, Others                       & Other Products                        \\
    \bottomrule
\end{tabular}
\end{table*}

\begin{table*}[]
    \scriptsize
    \centering
    \caption{Three nearest neighbors for selected stocks in the test set for the BERT+GCN+RC model}
    \label{tab:table4}
    \begin{tabular}{ccc}

                                        \toprule
                                        Company                           & TOPIX17                                       & TOPIX33                              \\
    \midrule
    \midrule

    Kintetsu Group   Holdings Co., Ltd. & Transportation \& Logistics               & Land Transportation                   \\
    \midrule
    Tobu Railway Co., Ltd.             & Transportation \& Logistics               & Land Transportation                   \\

    Kyushu   Railway Company          & Transportation \& Logistics               & Land Transportation                   \\

    Kobe Electric Railway Co., Ltd.      & Transportation \& Logistics               & Land Transportation                   \\
    \midrule
    \midrule
    Nippon Express Co., Ltd           & Transportation \& Logistics               & Land Transportation                   \\
    \midrule
    Inui Global Logistics Co., Ltd.    & Transportation \& Logistics               & Marine Transportation                 \\

    Kintetsu World Express, Inc       & Transportation \& Logistics               & Warehousing and Harbor Transportation \\

    Nippon Yusen Kabushiki Kaisha     & Transportation \& Logistics               & Marine Transportation                 \\
    \midrule
    \midrule
    Fujitsu General Limited           & Electric appliances \& Precision instruments & Electric Appliances                   \\
    \midrule
    Daihen Corporation                & Electric appliances \& Precision instruments & Electric Appliances                   \\

    Hokuetsu Industries Co., Ltd.      & Machinery                                   & Machinery                             \\

    Nissan Electric Co., Ltd.           & Electric appliances \& Precision instruments & Electric Appliances                   \\
    \midrule
    \midrule
    SBI Holdings, Inc.                 & Financial (except banks)                        & Securities and Commodities Futures    \\
    \midrule
    Mizuho Leasing Company, Limited   & Financial (except banks)                        & Other Financing Business              \\

    Money Partners Group Co., Ltd.      & Financial (except banks)                        & Securities and Commodities Futures    \\

    Toyo Securities Co., Ltd.           & Financial (except banks)                        & Securities and Commodities Futures   \\
    \bottomrule
\end{tabular}
\end{table*}

\begin{table*}[]
    \centering
    \caption{Ablation Study}
    \label{tab:table5}
    \begin{tabular}{cclccccccc}
    \toprule
        &                      &            &                & TOPIX17       &                &           &                & TOPIX33       &                \\
    \cmidrule(lr){4-6}
    \cmidrule(lr){8-10}
        BERT Train        & Graph Type           &             & MAP@5          & MAP@10         & MAP@50         &           & MAP@5          & MAP@10         & MAP@50         \\
        \midrule
        w/ train     & Directed             & BERT+GCN+RC    & ${\bf0.682}$ & 0.651          & 0.547          & ${\bf}$ & ${\bf0.606}$ & ${\bf0.587}$ & ${\bf0.528}$ \\
                         &                      & BERT+GAT+RC    & 0.679          & ${\bf0.657}$ & ${\bf0.561}$ & ${\bf}$ & 0.598          & 0.585          & 0.524          \\
                         &                      & RoBERTa+GCN+RC & 0.68           & 0.651          & 0.553          & ${\bf}$ & 0.599          & 0.579          & ${\bf0.528}$ \\
                         &                      & RoBERTa+GAT+RC & 0. 666         & 0.635          & 0.528          & ${\bf}$ & 0.581          & 0.555          & 0.495          \\
        \midrule
        w/ train     & Undirected           & BERT+GCN+RC    & 0.647          & 0.616          & 0.512          & ${\bf}$ & 0.568          & 0.536          & 0.485          \\
                         &                      & BERT+GAT+RC    & 0.612          & 0.569          & 0.468          & ${\bf}$ & 0.532          & 0.488          & 0.44           \\
                         &                      & RoBERTa+GCN+RC & 0.649          & 0.615          & 0.524          & ${\bf}$ & 0.552          & 0.522          & 0.471          \\
                         &                      & RoBERTa+GAT+RC & 0.663          & 0.627          & 0.523          & ${\bf}$ & 0.553          & 0.514          & 0.468          \\
        \midrule
        w/o train    & Directed             & BERT+GCN+RC    & 0.637          & 0.611          & 0.51           & ${\bf}$ & 0.567          & 0.547          & 0.481          \\
                         &                      & BERT+GAT+RC    & 0..66          & 0.623          & 0.501          & ${\bf}$ & 0.572          & 0.545          & 0.482          \\
                         &                      & RoBERTa+GCN+RC & 0.673          & 0.633          & 0.539          & ${\bf}$ & 0.597          & 0.56           & 0.507          \\
     &  & RoBERTa+GAT+RC & 0.661          & 0.624          & 0.509          & ${\bf}$ & 0.571          & 0.541          & 0.484         \\
     \bottomrule
    \end{tabular}
    \end{table*}

\begin{table*}[]
    \centering
    \small
    \caption{Results for the Thematic Fund Creation}
    \label{tab:table6}
    \begin{tabular}{lcccccccc}
        \toprule
        & LOGISTICS      & IT        &     & BIOTECHNOLOGY  & SEMICONDUCTOR  & & 5G             & POST-COVID     \\
    \cmidrule(r){2-3}
    \cmidrule(r){5-6}
    \cmidrule(r){8-9}

            Random Guess   & 0.031          & 0.033    &  ${\bf}$    & 0.033          & 0.033    &      & 0.037          & 0.035          \\
    \midrule
            BERT           & 0.308          & 0.227     &  ${\bf}$   & 0.251          & 0.067    &      & 0.062          & 0.118          \\
    RoBERTa        & 0.375          & 0.188      &  ${\bf}$  & ${\bf0.267}$ & 0.063    &      & 0.077          & 0.080          \\
    \midrule
    RoBERTa+GCN+RC & 0.313          & 0.294    &   ${\bf}$   & 0.220          & 0.125    &      & 0.065          & 0.080          \\
    RoBERTa+GAT+RC & ${\bf0.424}$ & 0.247     &  ${\bf}$   & 0.239          & 0.125    &      & 0.096          & 0.094          \\
    BERT+GCN+RC    & 0.388          & 0.208     &  ${\bf}$   & 0.204          & 0.114     &     & ${\bf0.105}$ & 0.125          \\
    BERT+GAT+RC    & 0.388          & ${\bf0.314}$ & ${\bf}$ & 0.227          & ${\bf0.145}$ & & 0.096          & ${\bf0.146}$ \\
    \bottomrule
    \end{tabular}
\end{table*}

The results for the related company information extraction task are reported in Table \ref{tab:table2}.
Tables \ref{tab:table3} and \ref{tab:table4} show the nearest neighbors for the selected stock using the BERT and BERT+GCN+RC models from the test data.

Figure \ref{fig:figure2} shows a visualization of the embeddings with TOPIX17 labels.
T-distributed stochastic neighbor embedding (t-sne) was applied to reduce the dimensions. 
The reduced embeddings were plotted and labeled with color using TOPIX17 labels. 
Figure \ref{fig:figure2a} shows the embeddings from the BERT model, whereas 
Figure \ref{fig:figure2b} shows the embeddings from the BERT+GCN+RC model. 
Table \ref{tab:table5} shows the results of the two types of ablation studies on the graph type and learning architecture.
Table \ref{tab:table6} shows the results of thematic fund creation for the six selected themes.

\section{Discussion}
\subsection{Related Company Information Extraction}
We compared the performance of the SETN with that of the baseline models in the task of related company information extraction. 
Table \ref{tab:table2} shows that the proposed modeling approach outperformed the baseline models. 
This indicates that jointly learning both textual and network information improves the performance of a model with regard to related company information extraction tasks.
In addition, adding a residual connection layer to the model improves its performance significantly. 
The residual connection layer enhances the model’s performance because it helps to equalize the weights of textual and network information by adding the output of BERT or RoBERTa, either of which encodes textual information, to the output of GCN or GAT, either of which encodes network information.

Figure \ref{fig:figure2} shows a visualization of the embeddings from the BERT and BERT+GCN+RC models with TOPIX17 labels. 
We can observe that the stocks in the same TOPIX17 sector tend to be grouped together more in the BERT+GCN+RC model than in the BERT model, which is consistent with the MAP@K evaluation in related company information extraction tasks.

Tables \ref{tab:table3} and \ref{tab:table4} show the three nearest neighbors for the selected stocks in the test data using the BERT and BERT+GCN+RC models.
The tables show that the three neighbors in the BERT+GCN+RC model generally have the same TOPIX17 and TOPIX33 labels as those in the BERT model.
The case of Kintetsu Group Holdings Co., Ltd provides an example where the TOPIX17 labels of the three nearest neighbors are correct in the BERT+GCN+RC model but incorrect in the BERT model. 
In the BERT model, the Kintetsu Department Store Co., Ltd. was chosen as one of the closest neighbors. 
Inspecting the textual data reveals that the business descriptions of both stocks are quite similar, which might be because Kintetsu Department Store Co., Ltd. is a subsidiary of Kintetsu Group Holdings Co., Ltd.
We consider the BERT model closely represents the two stocks because it is difficult to tell the difference between each stock's sector and industry in this example, as the model only learns from textual information. 
Conversely, the BERT+GCN+RC model learns from textual information and cross-company economic relationship information. 
Therefore, the embeddings become more informative for the stock sector and industry when the model is trained with both types of information. 
This example suggests that the proposed model can utilize both textual and network information.

Table \ref{tab:table5} presents the results of the two ablation studies.
The first ablation study investigated whether a directed graph model performed better than an undirected graph model. 
For the results, we compared the first and second sections in Table \ref{tab:table5}.
The results show that the model with a directed graph outperforms the model with an undirected graph for related company information extraction tasks. 
We observe that a directed graph model better captures cross-stock relationships because the economic relationship between stocks is often in a directed form, such as supply chain and transaction relationships. 

The second ablation study investigated how model performance improves by jointly training both the text embedding model and the graph embedding model. 
The results can be observed by comparing the first and third sections of Table \ref{tab:table5}. 
The result suggests that jointly training both text-embedding and graph embedding models outperforms separately encoding textual information with the transformer-based model and using it as an input to the GNN model. 
We observe that the embeddings learn both textual and network information more effectively by training both the transformer-based and GNN models in an end-to-end framework.

\subsection{Thematic Fund Creation}

The results of the thematic fund creation experiment reveal that our proposed models outperform baseline models in most settings. 
First, comparing the performances of the different models shows that our modeling approach outperformed the baseline and random guess models. 
This indicates that jointly learning both textual and network information improves the performance of models with regard to thematic fund creation tasks.
Second, a comparison of the performances of different theme groups revealed that the performances deteriorated from the first to the last group as the task gradually became different from the related company information extraction task. 
The first group, logistics and IT, had the highest performance among all groups.
An explanation for this is that the group’s task is similar to the related company information extraction task, and it has already been shown that the embeddings capture TOPIX17 and TOPIX33 information well. 
The tasks of the second group, biotechnology and semiconductor, can be seen as zero-shot learning of related company information extraction because the label is an unseen sector and industry label. The performance is not the highest but is close to that of the first group. We recall that during training, the embeddings learned the latent variables of the sector and industry associated with the stocks. This leads to competitive results, even for an unseen theme. 
The third group, 5G and Post-COVID, had the lowest performance, as the theme was completely different from the related company information extraction task, and it was not even about the company stocks associated industry or sector. However, our model outperformed the baseline and random-guess models.
Our analysis suggests that the embeddings obtain information about social situations, new technology, and other general concepts by learning from business descriptions and economic relationships captured by causal chains, although 
the scope of the information that the embeddings learned is not yet fully understood. Therefore, the properties of the embeddings must be analyzed thoroughly in future work.

\section{Conclusion}
We presented the SETN, a framework that jointly learns from textual and network information.
SETN learns textual information with transformer-based models, such as BERT and RoBERTa, and learns network information using GNN models, such as GCN and GAT.
We tested the validity of our proposed model on the related company information extraction task using MAP@K, which evaluated the extent to which the candidate models captured the information of the associated industrial sector.
Our proposed models, especially the BERT+GAT+RC model, outperformed the baseline models in related company information extraction tasks.
Furthermore, we conducted ablation studies to confirm the validity of using a directed graph to learn network information and the advantages of jointly training a transformer-based text embedding model and a GNN graph embedding model over training them separately.
Finally, we applied the embeddings obtained from our proposed model to a thematic fund creation task.
Our proposed model outperformed the baseline models, and the results showed that the embeddings captured the latent variables of stocks in the associated sector and industry.

We plan to extend this work to incorporate rich information about stocks, such as shareholding and supply chain relationships, in three ways. 
First, we shall widen the scope of the data to capture different types of network information. 
Second, we shall expand our model to heterogeneous graphs that can handle different types of nodes and edges. 
Third, we shall identify new evaluation metrics to quantify the extent to which the model captures  diverse network information.
\section*{Acknowledgment}

This work was supported by the Japan Science and Technology Future Program (JST-Mirai), Grant Number JPMJMI20B1, Japan.
This work was also supported by Daiwa Securities Group Inc.
\bibliographystyle{IEEEtran}
\bibliography{IEEE2022.bib}

\end{document}